\documentclass{article}

\PassOptionsToPackage{numbers,sort&compress}{natbib}
\usepackage[preprint]{neurips_2026}

\usepackage[utf8]{inputenc} %
\usepackage[T1]{fontenc}    %
\usepackage{hyperref}       %
\usepackage{url}            %
\usepackage{booktabs}       %
\usepackage{amsfonts}       %
\usepackage{nicefrac}       %
\usepackage{microtype}      %
\usepackage[dvipsnames]{xcolor} %
\usepackage{amsmath}
\usepackage{graphicx} 
\usepackage{multirow}
\usepackage{colortbl}
\usepackage{cleveref}
\usepackage{amssymb}
\usepackage{enumitem}
\usepackage{float}
\usepackage{bbm}
\usepackage{subcaption} %
\usepackage[most]{tcolorbox}    %

\usepackage{amsmath,amsfonts,bm}

\def\eqref#1{equation~\ref{#1}}

\def\1{\bm{1}}

\def\vepsilon{{\bm{\epsilon}}}

\def\vv{{\bm{v}}}

\def\vx{{\bm{x}}}

\DeclareMathAlphabet{\mathsfit}{\encodingdefault}{\sfdefault}{m}{sl}
\SetMathAlphabet{\mathsfit}{bold}{\encodingdefault}{\sfdefault}{bx}{n}

\crefname{equation}{Eq.}{Eqs.}
\crefname{table}{Tab.}{Tabs.}
\crefname{section}{Sec.}{Secs.}
\crefname{figure}{Fig.}{Figs.}

\title{Recency Forcing: Bridging the Long-Horizon Gap in Autoregressive Video Generation}

\newcommand{\equalcontrib}{$^{*}$}
\newcommand{\workdone}{$^{\dagger}$}
\newcommand{\QAIR}{$^{\dagger}$}

\author{%
  Tri Cao\equalcontrib \quad\quad Hung Nguyen\equalcontrib\workdone \quad\quad Phong Nguyen \quad\quad Khoi Nguyen \\[5pt]
  Qualcomm AI Research\QAIR \\[5pt]
  \texttt{\{tricc, hunnguy, phongnh, khoi\}@qti.qualcomm.com}
}

\definecolor{mydarkblue}{rgb}{0,0.08,1}
\definecolor{mydarkgreen}{rgb}{0.02,0.6,0.02}
\definecolor{myred}{rgb}{1.0,0.0,0.0}
\definecolor{myred2}{rgb}{0.7,0.1,0.1}
\definecolor{mydarkblue2}{rgb}{0.05,0.1,0.7}
\definecolor{mypurple}{rgb}{111,0,255}
\definecolor{mypurple2}{rgb}{111,0,111}
\definecolor{catgray}{gray}{0.92}

\newcommand{\myheading}[1]{\vspace{2mm}\noindent{\textbf{#1}}}

\begin{document}

\maketitle
\begingroup
\renewcommand{\thefootnote}{\fnsymbol{footnote}}
\footnotetext[1]{Equal contribution.}
\footnotetext[2]{Work primarily done while at Qualcomm AI Research.}
\footnotetext[3]{Qualcomm AI Research is an initiative of Qualcomm Technologies, Inc.}
\endgroup

\begin{abstract}

Autoregressive (AR) video generation degrades over long horizons due to an overlooked train-inference discrepancy we term \textit{KV eviction mismatch}: models train on short clips where all context frames reside in the KV cache, but at inference, memory constraints force distant frames to be evicted from the KV cache -- removing context the model was conditioned on. Rather than simulating eviction via context truncation -- which discards temporal information the model still needs and degrades motion coherence -- we keep the  context but while progressively reducing the influence of distant frames, making their eventual eviction negligible.
To guide this design, we introduce the \textbf{positional response} $R( \Delta, \, t_{\text{denoise}})$, a perturbation-based sensitivity measure revealing that context influence decays steeply with temporal distance and varies systematically across denoising steps. Motivated by this analysis, we propose \textbf{Recency Forcing}, which applies a non-positive, timestep-dependent bias, termed \textbf{Temporal Response Bias (TRB)}, on pre-softmax attention logits derived directly from $R$, closing the train–inference gap without modifying context length or training objectives. We further introduce \textbf{Biased Attention Reparameterization (BAR)}, an exact reformulation that moves the bias outside the softmax, making TRB a standard FlashAttention call at zero overhead. 
Recency Forcing operates in both \textit{training-free} mode and \textit{training-based} mode.
Experiments on VBench and VBench-Long demonstrate state-of-the-art long-horizon generation quality at no additional inference cost.
\end{abstract}

\section{Introduction}

\begin{figure}[t]
    \centering
    \includegraphics[width=.95\linewidth]{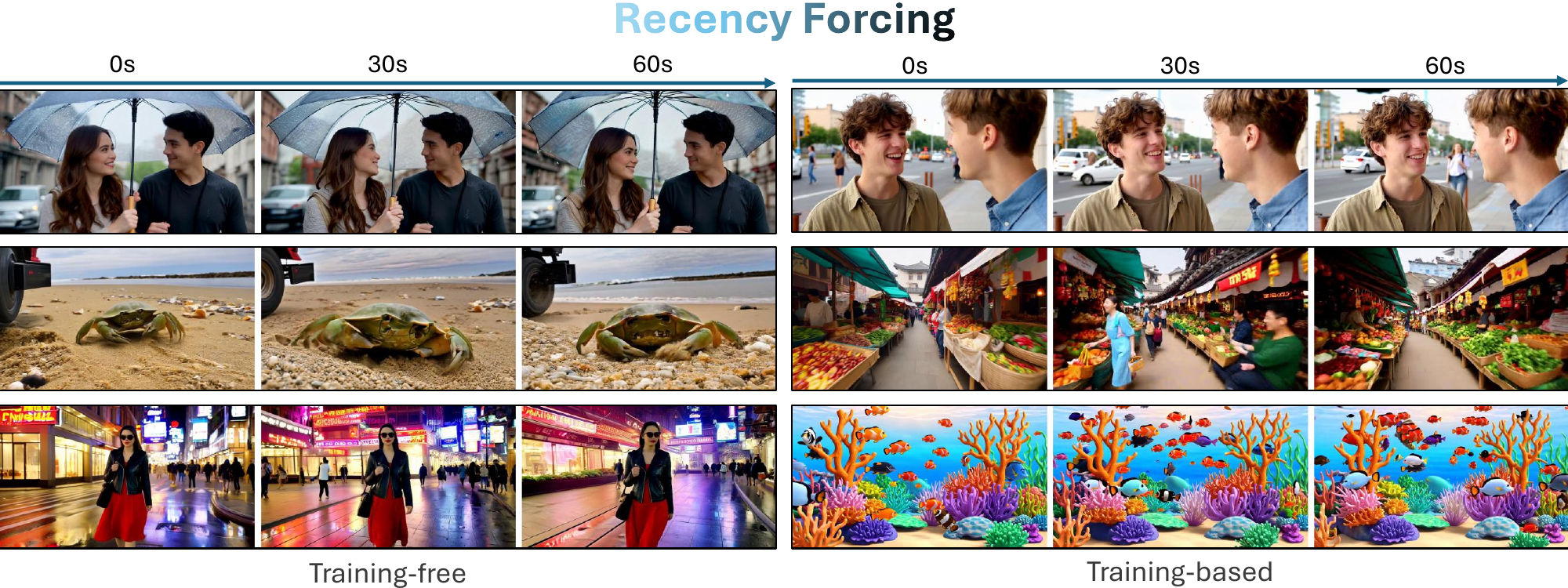}
    \caption{Our \textbf{Recency Forcing} enables high-quality video generation up to 60 seconds, far beyond the 5-second training horizon. Our method works as a plug-in (\textit{training-free}) module to improve the long-horizon performance of Self-Forcing, while further finetuning (\textit{training-based}) achieves state-of-the-art long-horizon video generation.}
    \label{fig:teaser}
    \vspace{-15pt}
\end{figure}

Autoregressive (AR) video generation predicts each frame from past context, enabling streaming and interactive applications with unbounded temporal rollouts, such as world modeling~\cite{mao2025yume, sun2025worldplay, hong2025relic}, game simulation~\cite{valevski2024diffusion, alonso2024diffusion, zhang2025matrix, yu2025gamefactory, tang2025hunyuan}, and robotics~\cite{li2025unified, yang2023learning}—settings where the bidirectional attention used by most state-of-the-art models~\cite{yang2024cogvideox, wan2025wan, kong2024hunyuanvideo} is fundamentally incompatible. Recent distillation-based methods (CausVid~\cite{yin2025causvid}, Self-Forcing~\cite{Huang2025SelfFB}, Causal Forcing~\cite{zhu2026causal}) enable fast few-step causal generation via distribution matching, but suffer from a key limitation: they are trained on short clips (e.g., 5 seconds, 21 latent frames in Wan2.1) and degrade rapidly when rolled out to long horizons.

Recent work extends these models to longer videos by adding an attention sink~\cite{xiao2023efficient} as a persistent anchor, then layering additional mechanisms on top. These efforts cluster into three families: \textbf{(1) KV-cache compression and pruning} controls memory growth by selectively pruning past context~\cite{mao2026packforcing, zhao2026relaxforcing, yi2025deep, chen2026past, yu2025malt}; \textbf{(2) RoPE-based methods} modify positional encoding for better length extrapolation~\cite{cui2026lol, yesiltepe2025infinity, li2026train, yang2026anchor, liu2025rolling}; \textbf{(3) correction-based methods} expose the model to its own degraded outputs or use RL feedback to stabilize rollout~\cite{Guo2025EndtoEndTF, Po2025BAggerBA, Xiang2026PathwiseTC, lu2025reward}.

We identify an aspect that prior work does not address: a train--test discrepancy we call \textbf{KV eviction mismatch}. During training, all context frames are accessible; at inference, memory constraints force distant frames to be evicted~\cite{yin2025causvid, Huang2025SelfFB}, removing context the model was trained to depend on. The most direct fix -- truncating context during training~\cite{lu2025reward, yang2025longlive} -- discards the temporal information needed for consistent motion, and we observe degraded motion fidelity as a result (empirically analyzed in~\cref{sec:context-decay} - \cref{fig:motivation_qualitative}). We instead keep the full context but progressively diminish the influence of distant frames so that by eviction time, their contribution is negligible.

The key question is \emph{what shape} this decay should take. We introduce the \textbf{positional response} $R(\Delta, \, t_{\text{denoise}})$, a perturbation-based sensitivity score measuring how the model's output changes when the context frame at temporal distance $\Delta$ from the current denoising frame is perturbed at denoising step $t_{\text{denoise}}$. The measured response reveals a two-axis structure: early timesteps focus sharply on nearby frames, while later timesteps draw information from a broader context window. Any fixed decay schedule cannot capture this, motivating a timestep-dependent design.

We propose \textbf{Recency Forcing}, whose core component is \textbf{Temporal Response Bias (TRB)}: an additive softmax bias decaying with temporal distance at a rate that varies with the denoising timestep -- steep early, gentle late -- mirroring $R$. Since distant frames already receive negligible weight at the eviction boundary, their removal becomes information-preserving, closing the train--test gap by design. Recency Forcing targets attention weighting, a layer none of the three prior families modifies, and composes cleanly with all of them. To make TRB practical, we introduce \textbf{Biased Attention Reparameterization (BAR)}, an exact reformulation that moves the bias outside the softmax, making TRB a standard FlashAttention~\cite{dao2023flashattention} call with no kernel modifications and zero inference overhead.

We evaluate on VBench~\cite{Huang2023VBenchCB} and VBench-Long~\cite{huang2025vbench++}, generating videos up to \textit{60} seconds—\textit{12$\times$} longer than the training horizon. The approach supports two modes: \emph{training-free}, requiring no modification to the base model, where our method serves as a plug-in enhancement to Self-Forcing~\cite{Huang2025SelfFB} and improves the VBench-Long quality score to \textit{82.63} with \textbf{no additional inference overhead}; and \emph{training-based}, requiring only $\sim$12 hours of DMD retraining, further improving the score to 84.02, achieving state-of-the-art performance and surpassing prior methods on several metrics. Illustrative results are shown in \cref{fig:teaser}.

Our contributions are as follows: 
\begin{itemize}[noitemsep, topsep=0pt]
    \item \textbf{Analysis.} The positional response $R(\Delta, t_{\text{denoise}})$, revealing a two-axis decay structure in context usage that no prior work has measured or exploited.
    \item \textbf{Method.} Recency Forcing with Temporal Response Bias (TRB), which closes the KV eviction mismatch by aligning attention weights with the measured $R$.
    \item \textbf{Implementation.} Biased Attention Reparameterization (BAR), an exact reformulation of TRB, enabling compatibility with FlashAttention kernel with zero inference overhead.
\end{itemize}

\section{Related Work}
\vspace{-0.2cm}
\myheading{Video generation models} have evolved from U-Net diffusion models~\cite{Blattmann2023StableVD, Guo2023AnimateDiffAY} to scalable Diffusion Transformers (DiT)~\cite{Peebles2022ScalableDM}, with recent models such as CogVideoX~\cite{yang2024cogvideox}, HunyuanVideo~\cite{kong2024hunyuanvideo}, Wan~\cite{wan2025wan}, Cosmos~\cite{Agarwal2025CosmosWF}, and LTX-Video~\cite{hacohen2024ltx} achieving high-quality synthesis via transformer architectures over spatiotemporal tokens. These models generate all frames jointly using bidirectional attention, serving as the backbone for many AR models. Among them, Wan~\cite{wan2025wan} has emerged as the dominant open-source base for AR video generation~\cite{yin2025causvid, Huang2025SelfFB, zhu2026causal}; we likewise build on Wan2.1.

\myheading{Autoregressive video generation.}
A growing line of work extends bidirectional video diffusion to autoregressive generation~\cite{henschel2025streamingt2v, li2024arlon, liu2024mardini, ren2025autoregressive, weng2024art, yuan2025lumos, Zhang2025GenerativePA}. Early AR methods such as NOVA~\cite{Deng2024AutoregressiveVG}, SkyReels~\cite{Chen2025SkyReelsV2IF}, and PyramidFlow~\cite{Lei2023PyramidFlowHD} generate chunk-by-chunk but require many sampling steps. More recent distillation-based methods achieve real-time generation: CausVid~\cite{yin2025causvid} distills a bidirectional DiT into a few-step causal generator; Self-Forcing~\cite{Huang2025SelfFB} reduces exposure bias by conditioning on the model's own outputs; Causal Forcing~\cite{zhu2026causal} further improves the ODE pretraining stage. All of these are trained on short clips ($\sim$5 seconds) and degrade over longer horizons --- the regime we target.

\myheading{Non-autoregressive long video generation.}
Most DiT-based models are limited to 5--10 second clips, and extending them is non-trivial. Hierarchical and sliding-window approaches such as NUWA-XL~\cite{Yin2023NUWAXLDO}, RIFLEx~\cite{Zhao2025RIFLExAF}, and FIFO-Diffusion~\cite{Kim2024FIFODiffusionGI} extend video length but do not exploit causal structure, making them less suitable for streaming settings.

\myheading{Autoregressive long video generation.}
The autoregressive paradigm is a natural fit for long-horizon generation, but AR models alone are insufficient: trained on short clips, they suffer from drift when extended, as observed in \cite{Huang2025SelfFB, zhu2026causal}. Recent work addresses this by extending distilled AR frameworks,  introducing an \textit{attention sink}~\cite{xiao2023efficient} as an anchor for long-term consistency, and further incorporating additional mechanisms. These efforts cluster into three families. \textbf{(1) KV-cache compression and pruning} controls memory growth by selectively discarding or compressing past context. DeepForcing~\cite{yi2025deep} uses participative compression to preserve important tokens while discarding redundant ones, reducing drift; related methods include PackForcing~\cite{mao2026packforcing}, RelaxForcing~\cite{zhao2026relaxforcing}, and MALT~\cite{yu2025malt}. \textbf{(2) RoPE-based methods} modify positional encoding for better length extrapolation, including $\infty$-RoPE~\cite{yesiltepe2025infinity}, Rolling Forcing~\cite{liu2025rolling}, LoL~\cite{cui2026lol}, and Anchor variants~\cite{yang2026anchor, li2026train}. \textbf{(3) Correction-based methods} expose the model to its own degraded outputs or use RL feedback to stabilize rollout. Self-Resampling~\cite{guo2025end} feeds the model its own generated errors, BAgger~\cite{Po2025BAggerBA} constructs corrective trajectories, Pathwise~\cite{Xiang2026PathwiseTC} aligns rollout trajectories with teacher distributions, and Self-Forcing++~\cite{cui2025selfforcing++} simulates long rollouts and optimizes with RL-based feedback~\cite{lu2025reward}.

These approaches improve long-horizon AR generation from different angles -- memory, positional generalization, and training robustness -- all aiming to extend context capacity. In contrast, our method addresses the train--inference mismatch by modulating attention to past frames based on temporal distance and denoising timestep.

\section{Method}
\label{sec:method}

\begin{figure}[t]
    \centering

    \begin{subfigure}[t]{0.9\linewidth}
        \centering
        \includegraphics[width=0.85\linewidth]{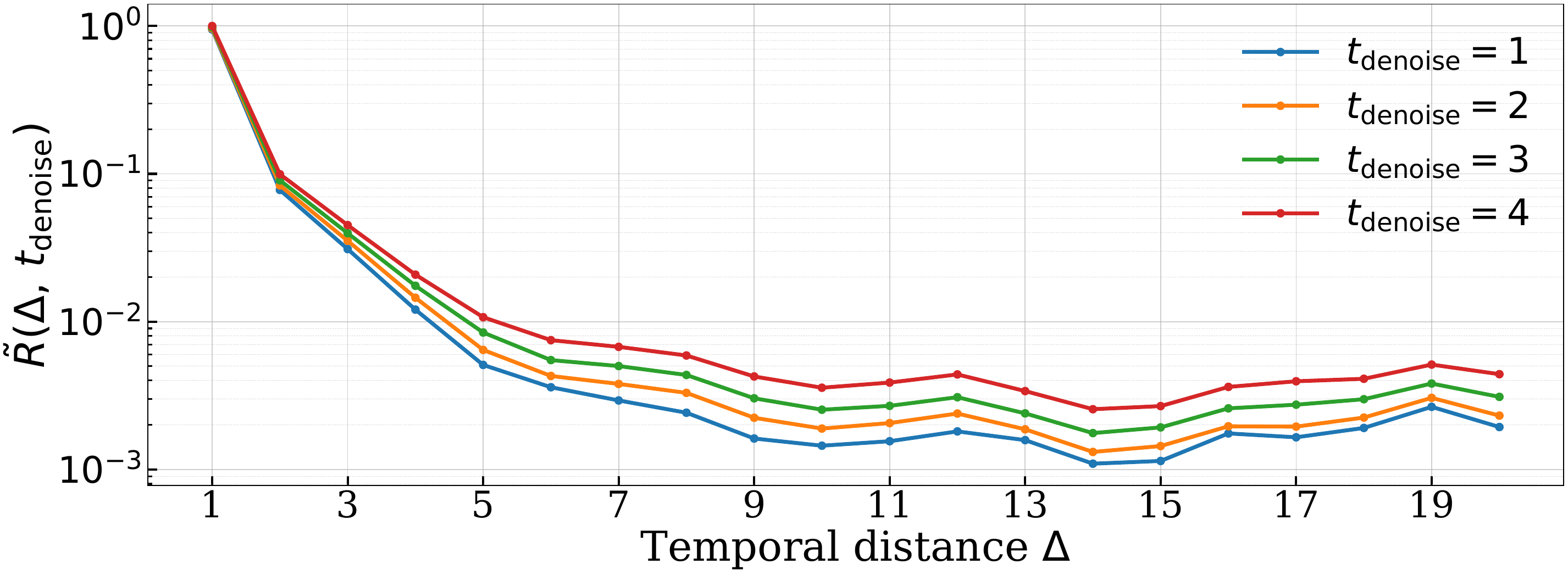}
        \caption{Normalized positional response $\tilde{R}(\Delta, \, t_{\text{denoise}})$ across denoising steps.}
        \label{fig:r_plot}
    \end{subfigure}

    \vspace{0.6em}

    \begin{subfigure}[t]{0.3\linewidth}
        \centering
        \includegraphics[width=\linewidth]{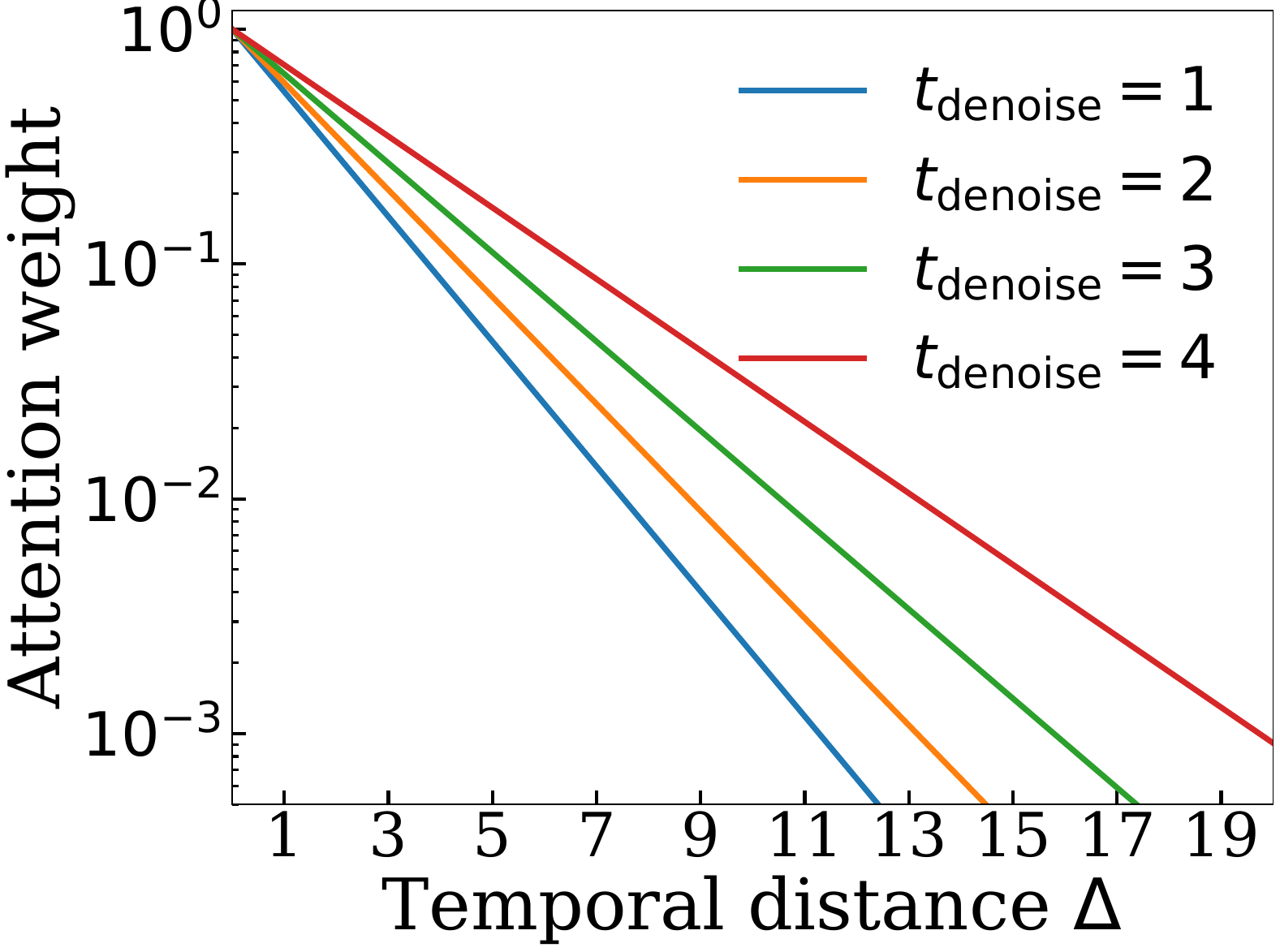}
        \caption{Linear decay.}
        \label{fig:linear_decay}
    \end{subfigure}
    \hfill
    \begin{subfigure}[t]{0.3\linewidth}
        \centering
        \includegraphics[width=\linewidth]{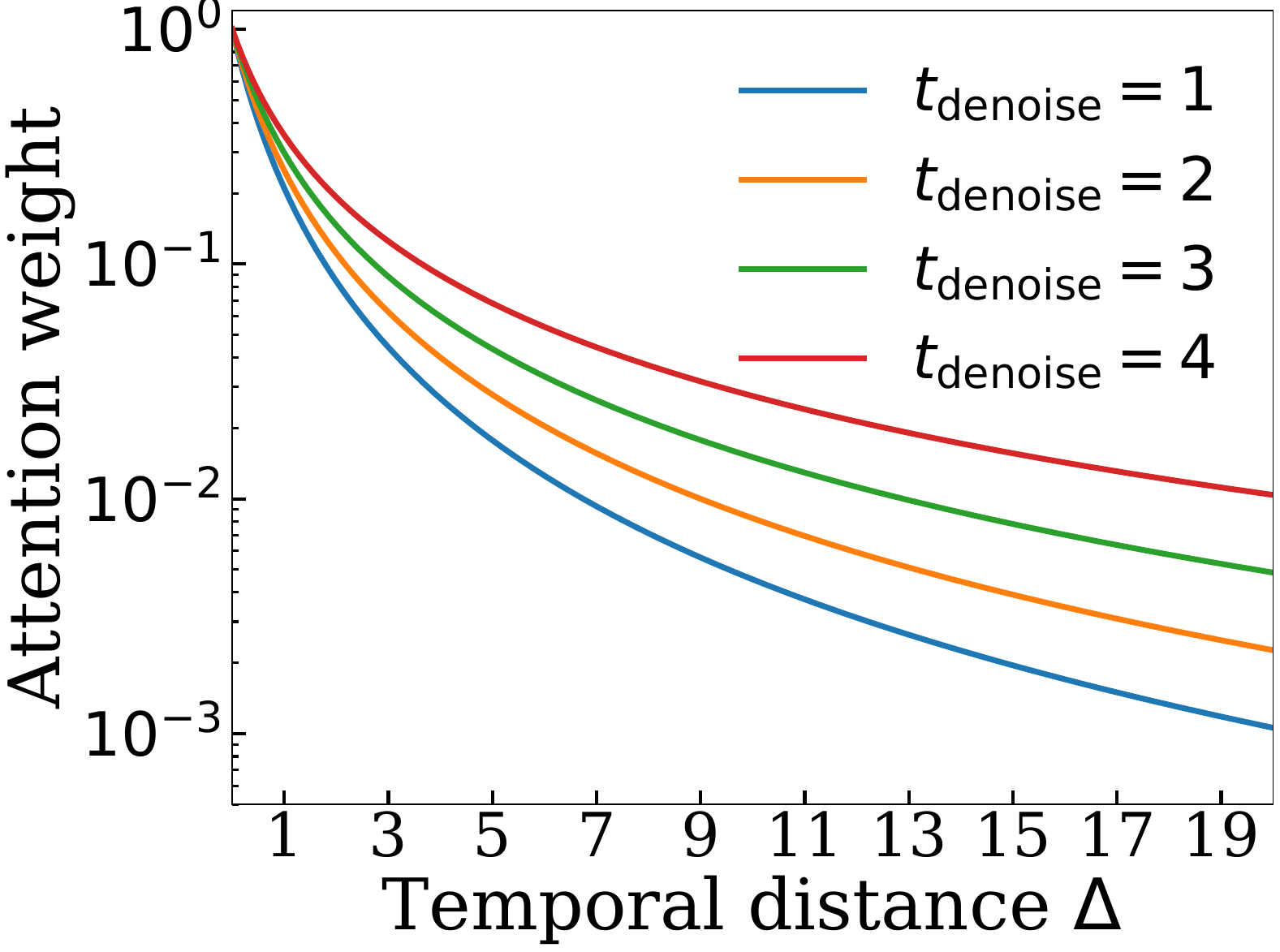}
        \caption{Logarithmic decay.}
        \label{fig:log_decay}
    \end{subfigure}
    \hfill
    \begin{subfigure}[t]{0.3\linewidth}
        \centering
        \includegraphics[width=\linewidth]{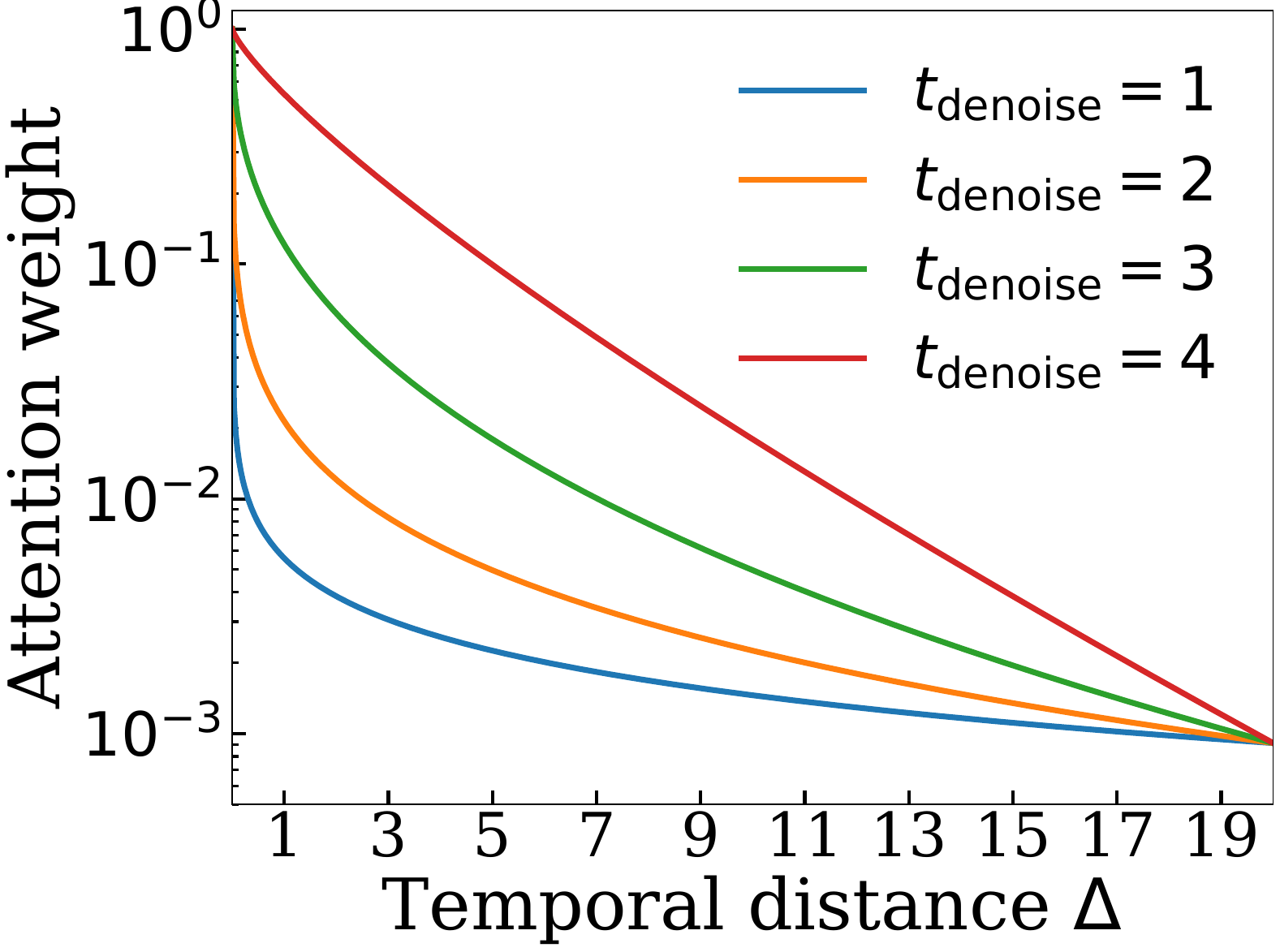}
        \caption{Power-law decay.}
        \label{fig:powerlaw_decay}
    \end{subfigure}

    \caption{
    \textbf{(a)} Normalized positional response $\tilde{R}(\Delta, \, t_{\text{denoise}})$ showing that context influence decays sharply with temporal distance and varies across denoising steps.
    \textbf{(b--d)} Candidate decay functions designed to follow the observed empirical pattern.
    }
    \label{fig:motivation}
    \vspace{-10pt}
\end{figure}

\subsection{Preliminaries}

\myheading{Flow matching.}
Flow matching~\cite{Lipman2022FlowMF, liu2022rectified, albergo2023stochastic} trains a velocity field $\vv_\theta(\vx_t, t)$ to transport a Gaussian prior to the data distribution. Following Wan2.1~\cite{wan2025wan} and its distillations~\cite{yin2025causvid, Huang2025SelfFB, zhu2026causal}, we use $t=1$ for pure noise and $t=0$ for clean data, with linear path $\vx_t = (1-t)\vx_0 + t\vepsilon$ and training loss
\begin{equation}
    \mathcal{L}_{\text{FM}}(\theta) = \mathbb{E}_{t, \vx_0, \vepsilon} \big\| \vv_\theta(\vx_t, t) - (\vepsilon - \vx_0) \big\|_2^2.
    \label{eq:fm_loss}
\end{equation}

\myheading{Autoregressive video generation.}
AR video models factorize the joint distribution as
\begin{equation}
p_\theta(\vx_0^{1:N}) = \prod_{i=1}^{N} p_\theta(\vx_0^i \mid \vx_0^{<i}),
\end{equation}
predicting each frame via a diffusion process conditioned on past generated frames $\vx_0^{<i}$ (\textit{context frames}). Generating a single frame requires many denoising steps (e.g., 50 in Wan2.1), making vanilla AR generation prohibitively slow for streaming use.

\subsection{Analysis: How Context Frames Influence Current Predictions}
\label{sec:analysis}

\myheading{Positional response.}
Considering a 4-step AR model, we propose a perturbation-based sensitivity score $R(\Delta, t_{\text{denoise}})$ that measures how strongly the model relies on the context frame at temporal distance $\Delta$ from the current denoising frame at denoising step $t_{\text{denoise}} \in \{1, 2, 3, 4 \}$. Here, $t_{\text{denoise}}=1$ denotes the highest-noise-level step, progressing to $t_{\text{denoise}}=4$ at the lowest-noise-level step.
\begin{equation}
    R(\Delta, \, t_{\text{denoise}})
    = \mathbb{E} \Big\|
        \vv_\theta\!\big(\vx_{t_{\text{denoise}}}^i \mid \vx_0^{<i}\big)
        - \vv_\theta\!\big(\vx_{t_{\text{denoise}}}^i \mid \delta_\Delta(\vx_0^{<i})\big)
    \Big\|^2,
    \label{eq:positional_response}
\end{equation}
where $\delta_\Delta(\vx_0^{<i})$ replaces the frame at position $\Delta$ with one from an alternative rollout of the same prompt (different random seed). We use replacement rather than additive noise to avoid introducing out-of-distribution perturbations. We normalize by the response at $\Delta = 1$ for cross-step comparison:
\begin{equation}
    \tilde{R}(\Delta, \, t_{\text{denoise}})
    = \frac{R(\Delta, \, t_{\text{denoise}})}{R(1, \, t_{\text{denoise}})}.
    \label{eq:normalized_response}
\end{equation}

\myheading{Observations.}
\cref{fig:r_plot} shows $\tilde{R}$ averaged over 30 prompts and 20 alternative contexts, measured on Causal Forcing~\cite{zhu2026causal}. Three patterns emerge consistently: \textbf{(1) monotonic ordering} -- curves are strictly ordered by $t_{\text{denoise}}$ at all distances; \textbf{(2) steep decay} -- $\tilde{R}$ drops by orders of magnitude within a few frames, regardless of denoising step; \textbf{(3) step-dependent long-range reach} -- responses at short distances are similar across steps, while responses at large $\Delta$ diverge substantially, indicating that denoising steps differ primarily in how they attend to distant context rather than in their reliance on nearby frames.

\begin{figure}[t]
    \centering
    \begin{minipage}[t]{0.62\linewidth}
        \centering
        \includegraphics[width=\linewidth]{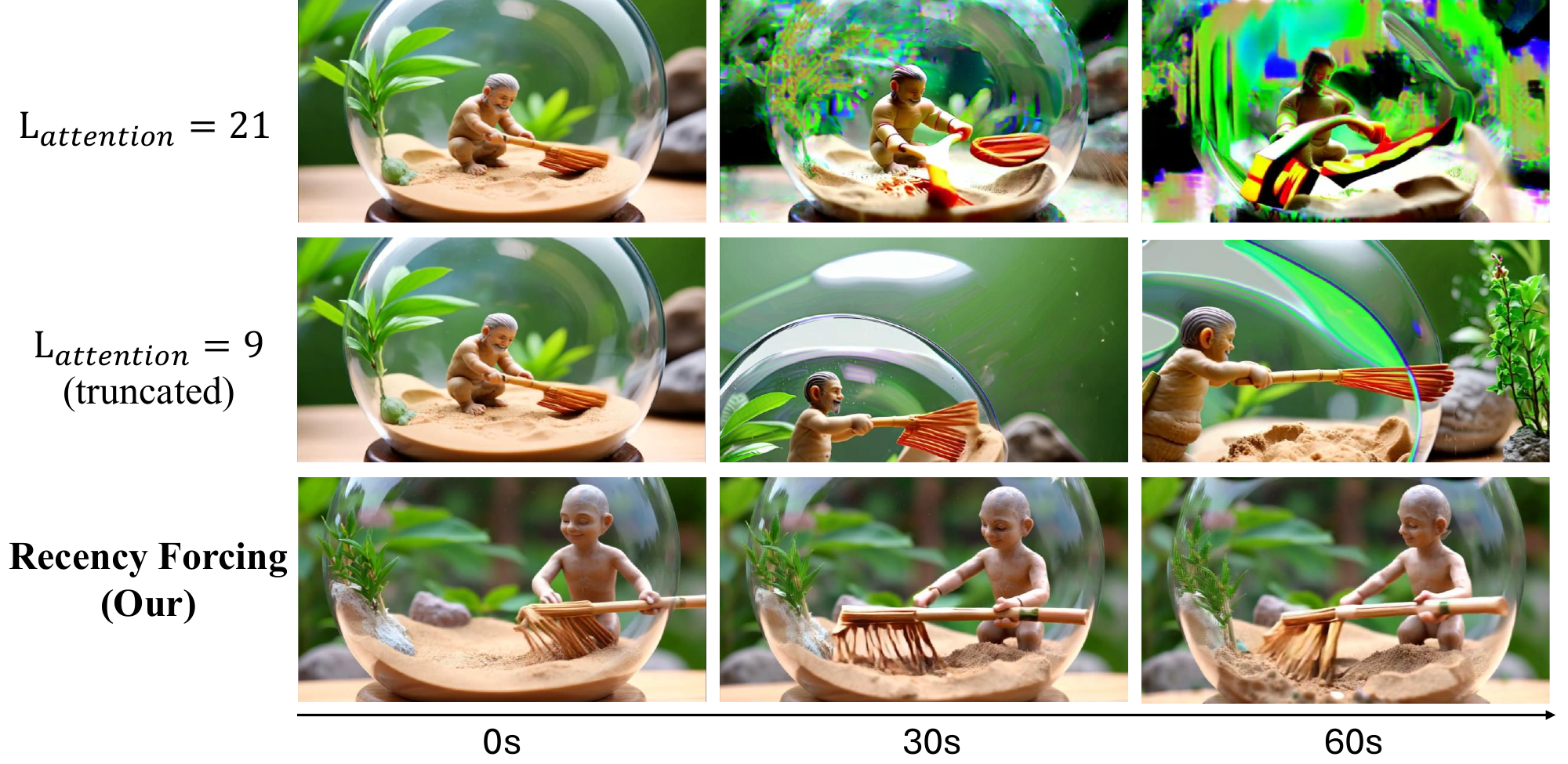}
        \caption{Qualitative comparison of context truncation ($L_{\text{attention}}=9$) and Recency Forcing over a 60-second rollout. Truncation produces unstable motion and scene changes (middle row), while Recency Forcing maintains consistent subject identity and visual quality throughout (bottom row).}
        \label{fig:motivation_qualitative}
    \end{minipage}
    \hfill
    \begin{minipage}[t]{0.34\linewidth}
        \centering
        \footnotesize
        \vspace{-60pt}
        \setlength{\tabcolsep}{2pt}
        \captionof{table}{Runtime comparison of FlexAttention with direct bias injection vs.\ our BAR-based FlashAttention. BAR achieves $11.76\times$ speedup with no approximation.}
        \label{tab:ttd_time_comparison}
        \vspace{4pt}
        \begin{tabular}{lcc}
            \toprule
            Method & Time (ms)$\downarrow$ & Speedup$\uparrow$ \\
            \midrule
            FlexAttn + bias & 70.34 & 1.00$\times$ \\
            Ours (FlashAttn) & 6.06 & \textbf{11.76$\times$} \\
            \bottomrule
        \end{tabular}
    \end{minipage}
\end{figure}

\myheading{Implications.}
A fixed attention pattern is suboptimal: any effective mechanism must \textbf{(a)} impose a temporal decay suppressing distant frames, and \textbf{(b)} modulate that decay with the denoising timestep. Context truncation satisfies (a) but abruptly removes frames the model still partially relies on, introducing motion discontinuities (as illustrated in the second row \cref{fig:motivation_qualitative}) and degrading performance (as reported in \cref{tab:context_decay_ablation}). We instead adopt continuous temporal attenuation, which suppresses distant-frame influence gradually while preserving the full context -- and by driving distant-frame weights to near zero before eviction, also closes the KV eviction mismatch.

We instantiate this as \textbf{Recency Forcing}, with \textbf{Temporal Response Bias (TRB)} -- a non-positive, timestep-dependent bias on pre-softmax logits derived from $\tilde{R}$ -- and \textbf{Biased Attention Reparameterization (BAR)}, an exact reformulation of TRB compatible with FlashAttention~\cite{dao2023flashattention} at zero overhead. Details follow in \cref{sec:temporal_response_bias,sec:bar}.

\subsection{Temporal Response Bias (TRB)}
\label{sec:temporal_response_bias}

We introduce TRB by adding a bias term to the scaled dot-product attention (SDPA) logits before the softmax, modulating how strongly the model attends to past frames based on both temporal distance $\Delta$ and denoising timestep $t_{\text{denoise}}$:
\begin{equation}
    \operatorname{SDPA}(Q, K, V;\, B) 
    = \operatorname{Softmax}\!\left( \frac{QK^\top}{\sqrt{d}} + B \right) V,
    \label{eq:biased_sdpa}
\end{equation}
where $B \in \mathbb{R}^{N \times N}$, $B_{i,j} = \text{bias}(\Delta_{i,j}, t_{\text{denoise}})$, and $\Delta_{i,j} = i - j$.

\begin{figure}[t]
    \centering
    \includegraphics[width=1.0\linewidth]{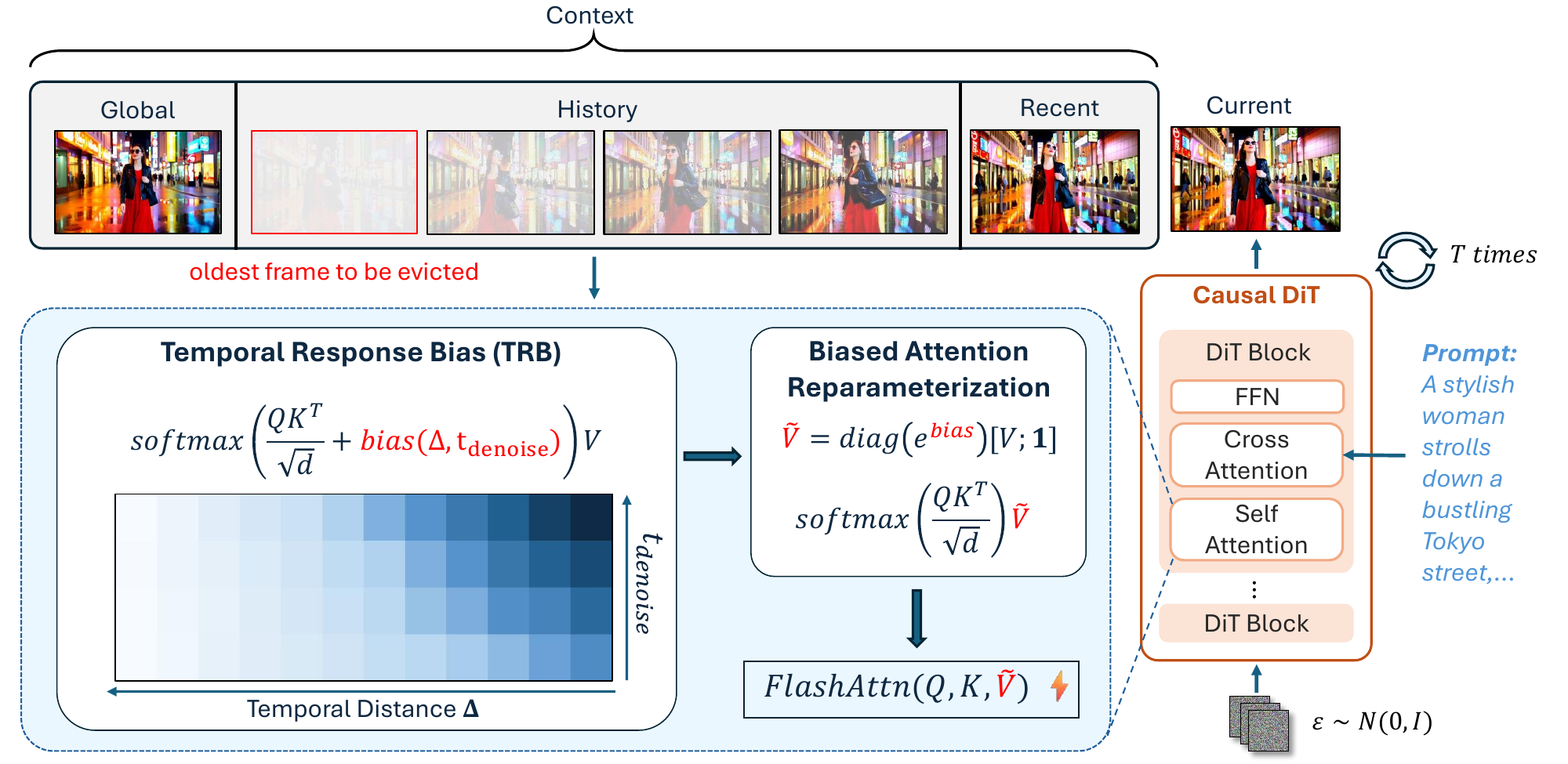}
    \caption{\textbf{Overview of Recency Forcing.} The context is partitioned into global sink frames ($L_{\text{global}}$), history frames ($L_{\text{history}}$), and recent frames ($L_{\text{recent}}$). Standard AR methods apply uniform attention over this window and evict distant frames at inference, creating a KV eviction mismatch. Recency Forcing addresses this by injecting \textbf{Temporal Response Bias (TRB)} into the attention logits, decaying the influence of history frames as a function of temporal distance $\Delta$ and denoising timestep $t_{\text{denoise}}$. While a direct implementation requires flexible kernels (e.g., FlexAttention), \textbf{Biased Attention Reparameterization (BAR)} provides an exact-equivalent reformulation that runs on standard FlashAttention at zero overhead.}
    \label{fig:ttd}
    \vspace{-10pt}
\end{figure}

\myheading{Context partition.} We partition context into three regions (\cref{fig:ttd}): \textbf{global} ($L_{\text{global}}$ initial frames as an attention sink), \textbf{recent} ($L_{\text{recent}}$ nearest frames for short-term motion), and \textbf{history} (intermediate frames). Including the current denoising frames $L_{\text{current}}$, the attention window size is defined as $L_{\text{attention}} = L_{\text{global}} + L_{\text{history}} + L_{\text{recent}} + L_{\text{current}}$. Recent and global contexts are kept unattenuated, as the positional response at $\Delta = 1$ remains near-maximal throughout denoising, while the global context is explicitly designated as an attention sink~\cite{liu2025rolling}. We thus apply TRB \textit{only} to the history context.

\myheading{Design principles.}
The bias satisfies three properties: \textbf{(1)} timestep-dependent strength -- stronger at early steps, weaker at later ones; \textbf{(2)} rapid decay with $\Delta$, matching the sharp drop in $\tilde{R}$; \textbf{(3)} non-positive values ($B_{i,j} \leq 0$), ensuring suppression of distant-frame attention rather than amplification.

\myheading{Candidate decay functions.}
We explore three families parameterized by $\alpha(t_{\text{denoise}}) > 0$ (decreasing function of $t_{\text{denoise}}$) and $\gamma(t_{\text{denoise}}) > 0$ (increasing function of $t_{\text{denoise}}$):
\begin{align}
    \text{bias}_{\text{lin}} &= -\,\alpha(t_{\text{denoise}}) \cdot (\Delta - L_{\text{recent}}), \label{eq:linear_decay} \\
    \text{bias}_{\text{log}} &= -\,\alpha(t_{\text{denoise}}) \cdot \log(1 + \Delta - L_{\text{recent}}), \label{eq:log_decay} \\
    \text{bias}_{\text{pow}} &= -\,\beta \cdot \tilde{\Delta}^{\,\gamma(t_{\text{denoise}})}, \quad \tilde{\Delta} = \frac{\Delta - L_{\text{recent}}}{L_{\text{history}}}, \label{eq:powerlaw_decay}
\end{align}
with $\beta > 0$ is a scale contant controlling the decay range. All three are illustrated in \cref{fig:linear_decay,fig:log_decay,fig:powerlaw_decay}.

Linear decay suppresses at a constant rate; logarithmic decay under-suppresses distant frames relative to $\tilde{R}$; power-law decay with $\gamma(t_{\text{denoise}})$ produces a convex profile matching the observed steep drop-off, concentrating suppression on distant frames. Ablations (\cref{sec:ablation}) confirm power-law performs best; we adopt it for all experiments.

\myheading{Timestep-dependent schedules.}
For our 4-step generator ($t_{\text{denoise}} \in \{1,2,3,4\}$), we use:
\begin{equation}
    \alpha(t_{\text{denoise}}) = \alpha_{\text{base}} - \frac{t_{\text{denoise}}}{4}, \qquad
    \gamma(t_{\text{denoise}}) = \gamma_{\text{base}} \cdot 2^{\,t_{\text{denoise}}},
    \label{eq:schedules}
\end{equation}
where $\alpha_{\text{base}}$ and $\gamma_{\text{base}}$ are the only free parameters. Decreasing $\alpha(t_{\text{denoise}})$ weakens suppression magnitude as denoising progresses; increasing $\gamma(t_{\text{denoise}})$ simultaneously sharpens the decay shape -- together reproducing the two-axis structure in $\tilde{R}$. In training-based mode, both base parameters are uniformly sampled during training; in training-free mode, they are set directly from $\tilde{R}$.

\subsection{Biased Attention Reparameterization (BAR)}
\label{sec:bar}
TRB requires injecting the bias $B$ into the pre-softmax logits. A direct implementation using flexible attention kernels such as FlexAttention~\cite{dong2024flex} supports this naturally but incurs non-trivial overhead and is incompatible with the fused softmax in FlashAttention~\cite{dao2023flashattention}, which does not expose intermediate logits. We resolve this with \textbf{Biased Attention Reparameterization (BAR)}: an exact algebraic reformulation that moves the bias outside the softmax, making TRB a standard FlashAttention call with a lightweight value augmentation.

For a single query frame $q_i$, since $B_{i,j}$ depends only on $\Delta_{i, \, j}$ and is independent of the content $(q_i, k_j)$, it can be factored as $w_j = \exp(B_{i,j}) \in (0,1]$, giving:
\begin{equation}
    o_i = \operatorname{Softmax}\!\left(\tfrac{q_i K^\top}{\sqrt{d}} + B_{i,:}\right) V
    = \frac{\sum_j \exp\!\left(\tfrac{q_i k_j^\top}{\sqrt{d}} + B_{i,j}\right) v_j}
           {\sum_j \exp\!\left(\tfrac{q_i k_j^\top}{\sqrt{d}} + B_{i,j}\right)}
    = \frac{\sum_j \exp\!\left(\tfrac{q_i k_j^\top}{\sqrt{d}}\right) w_j v_j}
           {\sum_j \exp\!\left(\tfrac{q_i k_j^\top}{\sqrt{d}}\right) w_j}.
    \label{eq:biased_attn_factor}
\end{equation}
Absorbing $w_j$ into an augmented value $\tilde{v}_j = w_j \cdot [v_j;\,1] \in \mathbb{R}^{d+1}$ where $[\cdot;\cdot]$ is concatenation and stacking into $\tilde{V} \in \mathbb{R}^{N\times(d+1)}$, this is equivalent to standard attention followed by a scalar normalization:
\begin{equation}
    \tilde{o}_i = \operatorname{Softmax}\!\left(\tfrac{q_i K^\top}{\sqrt{d}}\right)\tilde{V}, \qquad
    o_i = \tilde{o}_i[{:}d\,]\;/\;\tilde{o}_i[\,d\,].
    \label{eq:bar}
\end{equation}

This is algebraically identical to \cref{eq:biased_sdpa} with no approximation. 

\myheading{Efficiency.}
BAR requires three lightweight modifications to a standard attention call: (i) rescale each value vector $v_j$ by its decay weight $w_j = \exp(B_{i,j})$, (ii) append a scalar $w_j$ to $v_j$ to form $\tilde{v}_j$, and (iii) normalize the output by its final channel. Steps (i)--(ii) are $O(Nd)$ and (iii) is $O(N)$, both negligible relative to the $O(N^2 d)$ attention cost. The attention computation itself is an unmodified FlashAttention call, inheriting all of its memory efficiency and speed. Our BAR implementation achieves more than $11\times$ speedup over FlexAttention with direct logit modification, with no degradation in output quality. Detailed runtimes are included in \cref{tab:ttd_time_comparison}

\myheading{Training and inference modes.}
Recency Forcing supports two operating modes: \textit{training-free} and \textit{training-based}. In both modes, the decay weights $w_j = \exp(B_{i,j})$ are computed using the power-law function~\cref{eq:powerlaw_decay}. In the \textit{training-based} mode, the model is further fine-tuned during a short DMD retraining stage ($\sim$12 hours on 4 H100s), yielding additional gains. Importantly, BAR introduces no additional computational cost at inference in either mode.

\section{Experiments}

\myheading{Implementation details.}
We build Recency Forcing on Wan2.1-T2V-1.3B~\cite{wan2025wan}, replacing standard SDPA with BAR-based TRB in all self-attention layers. The \textit{training-free} variant is applied directly to chunk-wise Self-Forcing~\cite{Huang2025SelfFB}, while the \textit{training-based} variant fine-tunes only the DMD stage of Causal Forcing~\cite{zhu2026causal}. We use $L_{\text{attention}} = 21$ and $L_{\text{global}} = L_{\text{recent}} = 3$. Additional training and implementation details are provided in Appendix-~\cref{app:implementation_detail}.

\myheading{Evaluation.}
We evaluate on VBench~\cite{Huang2023VBenchCB} for short-video quality and VBench-Long~\cite{huang2025vbench++} for long-horizon consistency, using 200 MovieGen~\cite{polyak2024movie} prompts following Rolling Forcing~\cite{liu2025rolling}, with a disjoint 50-prompt set for ablations. All videos are 60 seconds at 16\,fps, $832{\times}480$. We additionally conduct a user study with 23 participants across four aspects: color consistency, motion dynamics, subject consistency, and overall quality. Details are in Appendix-\cref{app:user_study}.

\myheading{Baselines.}
We compare against recent autoregressive video generation methods across two settings.
For \textit{short video generation} (VBench), we include SkyReelsV2~\cite{Chen2025SkyReelsV2IF}, MAGI-1~\cite{ai2025MAGI1AV}, NOVA~\cite{Deng2024AutoregressiveVG}, Pyramid Flow~\cite{Lei2023PyramidFlowHD}, CausVid~\cite{yin2025causvid}, Self-Forcing~\cite{Huang2025SelfFB}, LongLive~\cite{yang2025longlive}, Causal Forcing~\cite{zhu2026causal}, and Rolling Forcing~\cite{liu2025rolling}.
For \textit{long video generation} (VBench-Long), we compare against both training-free methods -- Infinity RoPE~\cite{yesiltepe2025infinity} and Deep Forcing~\cite{yi2025deep} -- and training-based AR methods --- Self-Forcing~\cite{Huang2025SelfFB}, Causal Forcing~\cite{zhu2026causal}, Rolling Forcing~\cite{liu2025rolling}, and LongLive~\cite{yang2025longlive}.

\subsection{Comparison with prior approaches}

\myheading{Quantitative results.}
\cref{tab:main_result_vbench} reports VBench results for short video quality, and \cref{tab:main_result_vbench_long} reports VBench-Long results for long-horizon generation. We highlight three takeaways.
\textbf{First, short video quality is preserved.} Our method achieves the best overall score on VBench, demonstrating that integrating TRB introduces no degradation in short-video quality -- the bias has negligible effect when all context frames fit within the training window.
\textbf{Second, long-horizon stability is substantially improved.} On VBench-Long, our method achieves the highest dynamic degree (75.48) while maintaining strong imaging quality (70.50) and subject consistency (97.68). This indicates that TRB effectively mitigates long-horizon drift without sacrificing visual fidelity or motion naturalness.
\textbf{Third, the training-free variant is competitive.} Among training-free methods, our approach achieves the best overall VBench-Long score, confirming that the positional response curves $\tilde{R}$ provide a reliable prior for setting the decay schedule without any additional training.

\begin{figure}[t]
    \centering
    \includegraphics[width=.95\linewidth]{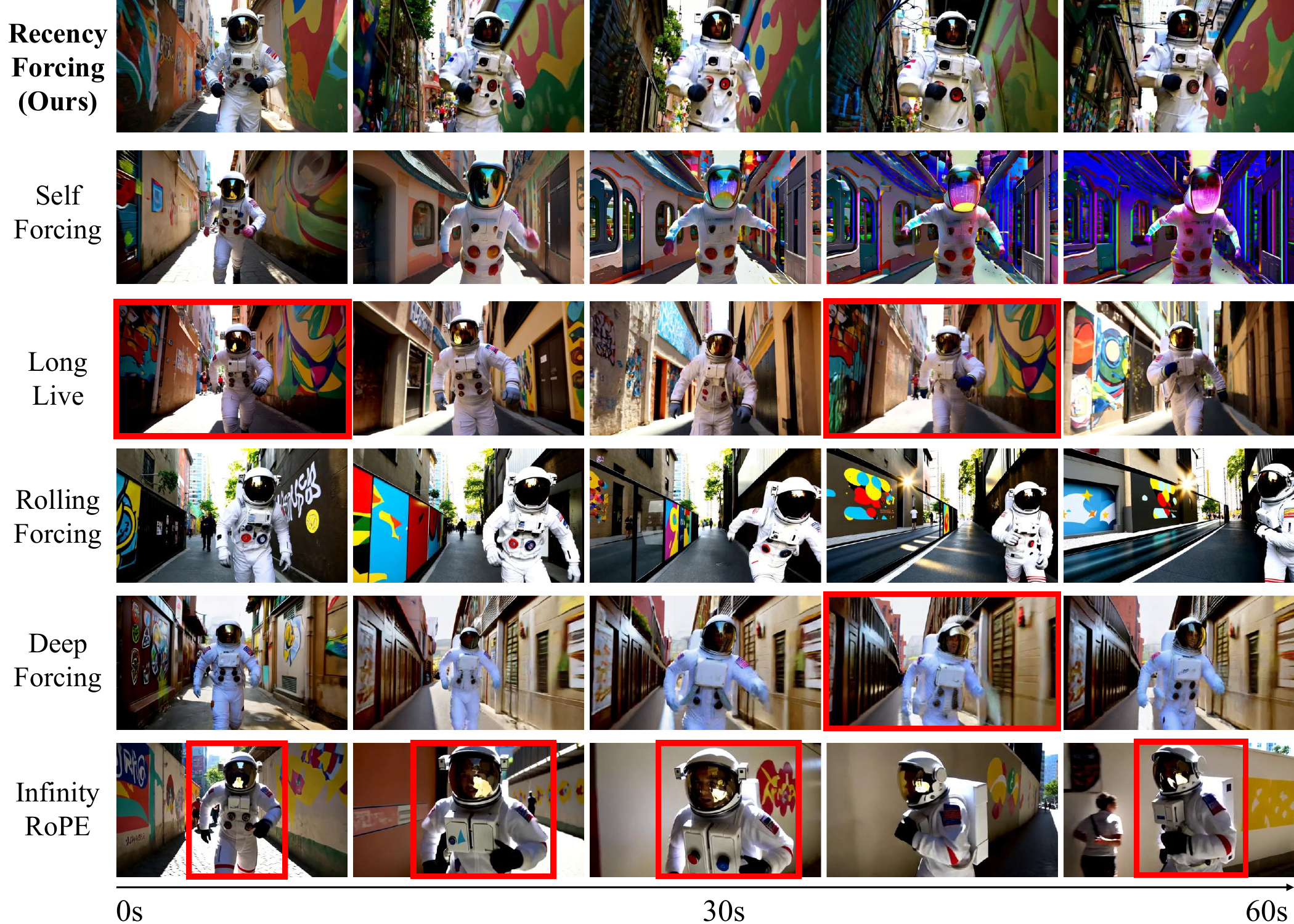}

    \caption{\textbf{Qualitative comparison} on 60-second videos. Self-Forcing suffers from color drift, LongLive produces scene repetition (\textcolor{red}{red boxes}), DeepForcing introduces blur, and Infinity-RoPE loses subject consistency. Recency Forcing preserves visual fidelity and motion throughout. For better visualization, we refer the reader to Appendix~\cref{app:additional_results} in the supplementary material.}
    \label{fig:qualitative}
    \vspace{-10pt}
\end{figure}

\myheading{Qualitative results.}
\cref{fig:qualitative} shows 60-second video generations across methods. Competing methods exhibit three main failure modes.
\textbf{Self-Forcing} shows progressive drift in color, texture, and sharpness, leading to quality degredation.
\textbf{LongLive} reduces drift but introduces repetition due to its global sink mechanism (\textcolor{red}{red boxes} in \cref{fig:qualitative}).
\textbf{Deep-Forcing} reduces drift but introduces compression artifacts such as motion blur.
\textbf{Infinity-RoPE} mitigates drift but suffers from gradual subject inconsistency over time.
In contrast, our method maintains consistent visual quality and motion dynamics over the full 60 seconds without noticeable artifacts or inconsistency.

\begin{table}[t]
  \small
  \setlength{\tabcolsep}{5pt}
  \caption{
    \textbf{Comparison on VBench on generated short (5-second) videos}. Higher is better.
  }
  \label{tab:main_result_vbench}
  \centering
  \begin{tabular}{lccc}
    \toprule
    \textbf{Model} & \textbf{Total Score } & \textbf{Quality Score}  & \textbf{Semantic Score} \\
    \midrule
    \rowcolor{catgray}
    \multicolumn{4}{l}{\textit{Autoregressive Video Diffusion Models}}\\
    SkyReels-V2~\cite{Chen2025SkyReelsV2IF}
                        & 82.67 & 84.70 & 74.53 \\
    MAGI-1~\cite{ai2025MAGI1AV}
                        & 79.18 & 82.04 & 67.74 \\
    NOVA~\cite{Deng2024AutoregressiveVG}
                        & 80.12 & 80.39 & 79.05 \\
    Pyramid Flow~\cite{Lei2023PyramidFlowHD}
                        & 81.72 & 84.74 & 69.62 \\
    \midrule
    \rowcolor{catgray}
    \multicolumn{4}{l}{\textit{Distilled Autoregressive Video Models}}\\
    CausVid~\cite{yin2025causvid}$^{*}$
                        & 81.20 & 84.05 & 69.80 \\
    Self Forcing~\cite{Huang2025SelfFB}        & 84.31 & 85.07 & \underline{81.28} \\
    Causal Forcing~\cite{zhu2026causal}      & 84.04 & 84.59 & \textbf{81.84} \\
    Rolling Forcing~\cite{liu2025rolling}     & 81.22 & 84.08 & 69.78 \\
    LONGLIVE~\cite{yang2025longlive}            & \underline{84.87} & \textbf{86.97} & 76.47 \\
    \midrule
    \textbf{Recency Forcing (Ours)}
                        & \textbf{85.08} & \underline{86.20} & 80.59 \\
    \bottomrule
  \end{tabular}
  \vspace{2pt}

  {\footnotesize\raggedright
  $^{*}$ We compare with the official implementation of CausVid using the same base model (Wan-1.3B).
  \par}
  \vspace{-10pt}
\end{table}

\begin{table}[t]
  \centering
  \setlength{\tabcolsep}{2.2pt}
  \caption{
   \textbf{Results on VBench-Long~\cite{huang2025vbench++} evaluated on 60-second videos.} Higher is better.
  }
  \label{tab:main_result_vbench_long}

  \resizebox{\linewidth}{!}{
  \begin{tabular}{lcccccccc}
    \toprule
    \textbf{Model}
    & \shortstack{\textbf{Quality}}
    & \shortstack{\textbf{Dynamic}}
    & \shortstack{\textbf{Motion}\\\textbf{Smoothness}}
    & \shortstack{\textbf{Temporal}\\\textbf{Flickering}}
    & \shortstack{\textbf{Imaging}}
    & \shortstack{\textbf{Aesthetic}}
    & \shortstack{\textbf{Subject}\\\textbf{Consistency}}
    & \shortstack{\textbf{Background}\\\textbf{Consistency}} \\
    \midrule

    \rowcolor{catgray}
    \multicolumn{9}{c}{\textit{Training-free}}\\

    Infinity RoPE~\cite{yesiltepe2025infinity}
      & 82.30 & 56.22 & 98.53 & 97.41 & 67.14 & 59.20 & 97.31 & 96.15 \\

    Deep Forcing~\cite{yi2025deep}
      & 82.14 & 56.01 & 98.10 & 96.88 & 68.22 & 59.85 & 97.35 & 96.30 \\

    \midrule

    \textbf{Recency Forcing (Ours)}
      & \underline{82.63} & 56.27 & 98.23 & 96.98 & 68.94 & 60.48 & \textbf{97.81} & \textbf{96.64} \\

    \midrule

    \rowcolor{catgray}
    \multicolumn{9}{c}{\textit{Training-based}}\\

    Self Forcing~\cite{Huang2025SelfFB}
      & 80.11 & 34.50 & 98.48 & \textbf{97.77} & 65.93 & 56.55 & 96.97 & 96.24 \\

    Causal Forcing~\cite{zhu2026causal}
      & 78.98 & \underline{65.19} & 96.94 & 95.21 & 63.44 & 51.35 & 95.39 & 95.51 \\

    Rolling Forcing~\cite{liu2025rolling}
      & 81.56 & 34.68 & \underline{98.71} & \underline{97.62} & \underline{70.18} & 59.93 & \underline{97.74} & \underline{96.54} \\

    LongLive~\cite{yang2025longlive}
      & 81.98 & 42.85 & \textbf{98.75} & 97.61 & 68.71 & \textbf{61.48} & 97.09 & 95.99 \\

    \midrule

    \textbf{Recency Forcing (Ours)}
      & \textbf{84.02} & \textbf{75.48} & 98.00 & 96.32 & \textbf{70.50} & \underline{61.31} & 97.68 & 96.49 \\

    \bottomrule
  \end{tabular}
  }

  \vspace{-0.5em}
\end{table}

\subsection{Ablation Studies}
\label{sec:ablation}

We conduct ablation studies on a held-out set of 50 prompts as the validation set, evaluating on VBench-Long with 60-second videos. All metrics are \textbf{higher-is-better}. Unless stated otherwise, ablations use the training-based variant with power-law decay.

\myheading{Context truncation vs.\ temporal decay.}
\label{sec:context-decay}
We first compare the two design directions in \cref{sec:analysis} for mitigating KV eviction mismatch in \cref{fig:motivation_qualitative} and \cref{tab:context_decay_ablation}. Context truncation (9-frame context) partially reduces the mismatch but removes useful temporal context, causing unnatural motion and unwanted scene change. This increases the dynamic degree (88.58) while harming the overall quality. In contrast, our temporal decay suppresses distant-frame influence smoothly, achieving better VBench-Long performance than both truncation and Causal Forcing.

\myheading{Choice of bias function.}
We compare linear, logarithmic, and power-law decay forms (\cref{eq:linear_decay,eq:log_decay,eq:powerlaw_decay}). As shown in \cref{tab:bias_ablation_long}, power-law decay performs best overall, consistent with its convex profile better matching the steep drop-off observed in the positional response curves $\tilde{R}$. Linear and logarithmic decay both under-suppress distant frames relative to the empirical response, leading to residual drift. We adopt power-law decay for all subsequent experiments.

\myheading{Effect of decay strength $\gamma_{\text{base}}$.}
We vary $\gamma_{\text{base}}$ at inference time to study decay aggressiveness. As shown in \cref{tab:gamma_ablation_long}, small values insufficiently suppress distant frames, while large values over-suppress history and harm motion coherence. We find $\gamma_{\text{base}} = 0.1$ provides the best trade-off.

\myheading{Effect of recent context length.}
We vary $L_{\text{recent}} \in \{0, 3, 6\}$ to study the effect of recent context length. As shown in \cref{tab:recent_ablation_long}, removing recent context entirely ($L_{\text{recent}} = 0$) significantly degrades motion quality, showing that nearby frames provide critical motion cues. Larger $L_{\text{recent}}$ improves motion stability with diminishing returns beyond a small window. We use $L_{\text{recent}} = 3$ by default.

\begin{table*}[!t]
    \centering
    \small
    \setlength{\tabcolsep}{5pt}
    \vspace{10pt}
    \begin{minipage}[t]{0.49\linewidth}
        \centering
        \caption{Context truncation vs.\ TRB decay.
        Row 1 ($L_{\text{attention}}=21$, no decay) is the Causal Forcing baseline. Truncation ($L_{\text{attention}}=9$) inflates dynamic degree at the cost of overall quality; TRB achieves the best balance.
        }
        \vspace{-5pt}
        \begin{tabular}{cc|ccc}
            \toprule
            $L_{\text{attention}}$ & \textbf{Decay}
            & \textbf{Quality} 
            & \textbf{Imaging}
            & \textbf{Dynamic}  \\
            \midrule
            21 & none & 80.56 & 65.09 & 78.32 \\
            9 & none & \underline{81.10} & \underline{66.40} & \textbf{88.58} \\
            21 & TRB & \textbf{84.17} & \textbf{69.05} & \underline{81.55} \\
            \bottomrule
            \vspace{-15px}
        \end{tabular}
        \label{tab:context_decay_ablation}
    \end{minipage}
    \hfill
    \begin{minipage}[t]{0.47\linewidth}
        \centering
        \caption{Effect of bias function. 
        Power-law decay best matches the steep drop-off in the positional response $\tilde{R}$, yielding the highest quality and dynamic degree.
        }
        \vspace{-5pt}
        \resizebox{\linewidth}{!}{
        \begin{tabular}{lccc}
            \toprule
            Bias($\Delta,t_{\text{denoise}}$)
            & \textbf{Quality}
            & \textbf{Imaging}
            & \textbf{Dynamic} \\
            \midrule
            Linear       & \underline{83.26} & 68.90 & 66.06 \\
            Log          & 83.16 & \textbf{69.42} & \underline{72.06} \\
            Power-law    & \textbf{84.17} & \underline{69.05} & \textbf{81.55} \\
            \bottomrule
        \end{tabular}
        }
        
        \label{tab:bias_ablation_long}
    \end{minipage}
    \vspace{-10pt}
\end{table*}

\begin{table*}[!t]
    \centering
    \small
    \setlength{\tabcolsep}{5pt}
    \vspace{10pt}
    \begin{minipage}[t]{0.45\linewidth}
        \centering
        \caption{Effect of decay strength $\gamma_{\text{base}}$.}
        \vspace{-5pt}
        \label{tab:gamma_ablation_long}
        \begin{tabular}{cccc}
            \toprule
            $\gamma_{\text{base}}$
            & \textbf{Quality} & \textbf{Imaging}  & \textbf{Dynamic}  \\
            \midrule
            0.01 & \textbf{84.36} & \textbf{69.63} & 79.42 \\
            0.1  & \underline{84.17} & \underline{69.05} & 81.55 \\
            0.5  & 82.91 & 67.05 & \underline{81.94} \\
            1.0  & 82.61 & 65.98 & \textbf{83.93} \\
            \bottomrule
        \end{tabular}
    \end{minipage}
    \hfill
    \begin{minipage}[t]{0.48\linewidth}
        \centering
        \caption{Effect of recent context size $L_{recent}$.}
        \vspace{-5pt}
        \label{tab:recent_ablation_long}
        \begin{tabular}{cccc}
            \toprule
            $L_{\text{recent}}$
            & \textbf{Quality}  
            & \textbf{Imaging}
            & \textbf{Dynamic}\\
            \midrule
            0 & \underline{82.80} & \textbf{69.86} & 40.90 \\
            3 & \textbf{84.17} & {69.05} & \textbf{81.55} \\
            6 & 82.14 & \underline{69.51} & \underline{49.03} \\
            \bottomrule
        \end{tabular}
    \end{minipage}
    
    \label{tab:ablation_combined}
\vspace{-5pt}
\end{table*}

\section{Conclusion}
\label{sec:conclusion}

We propose \textbf{Recency Forcing}, whose core component is \textbf{Temporal Response Bias (TRB)}: an additive softmax bias decaying with temporal distance at a rate that varies with the denoising timestep -- steep early, gentle late -- mirroring $R$. Since distant frames already receive negligible weight at the eviction boundary, their removal becomes information-preserving, closing the train--test gap by design. Recency Forcing targets attention weighting, a layer none of the three prior families modifies, and composes cleanly with all of them. To make TRB practical, we introduce \textbf{Biased Attention Reparameterization (BAR)}, an exact reformulation that moves the bias outside the softmax, making TRB a standard FlashAttention~\cite{dao2023flashattention} call with no kernel modifications and zero inference overhead.

\newpage

{
  \small
  \bibliographystyle{plain}
  \bibliography{main}
}

\clearpage
\appendix

\section{Additional qualitative results}
\label{app:additional_results}
We provide additional qualitative and comparison results on our \textit{Project Web Page}.

\section{Implementation Details}
\label{app:implementation_detail}

We implement Recency Forcing on top of a 1.3B autoregressive video diffusion model following Causal Forcing~\cite{zhu2026causal}, built on the Wan2.1-T2V-1.3B backbone~\cite{wan2025wan}. We replace standard SDPA with BAR-based TRB in all self-attention layers. The model generates videos at a resolution of $832 \times 480$, $L_{\text{attention}} = 21$ latent frames, $L_{\text{global}} = L_{\text{recent}} = L_{\text{current}} = 3$ latent frames, with TRB applied to the remaining history frames.

\myheading{Analysis.} The positional response $R(\Delta, \, t_{\text{dempose}})$ analysis is conducted on both chunk-wise and frame-wise Causal Forcing~\cite{zhu2026causal}, only frame-wise version is reported in~\cref{sec:analysis} for conceptual clarity.

\myheading{Training.}
Our method can be directly applied to pretrained distilled autoregressive video models~\cite{Huang2025SelfFB,zhu2026causal}. For the training-based variant, we initialize from a pretrained causal ODE checkpoint of Causal Forcing~\cite{zhu2026causal} and fine-tune only the DMD stage with TRB following the Self-Forcing~\cite{Huang2025SelfFB} training protocol. Training is conducted on short clips (5 seconds) sampled from a filtered and LLM-augmented version of VidProM~\cite{wang2024vidprom} for 3{,}500 steps with batch size 4 on H100 GPUs. During training, we sample $\alpha_{\text{base}} \sim \mathcal{U}(0, 2)$ and $\gamma_{\text{base}} \sim \mathcal{U}(0, 1)$, and set $\beta = L_{\text{history}}$.

\myheading{Inference.}
During inference, KV-cache eviction is applied as in prior work, while TRB adaptively reduces reliance on evicted frames. All experiments use chunk-wise Causal Forcing~\cite{zhu2026causal} as the base model, with a chunk size of 3 latent frames, temporal distance $\Delta$ is calculated between chunks. For the training-free setting, we apply our method to Self-Forcing~\cite{Huang2025SelfFB} to ensure a fair comparison with other training-free methods (e.g., Deep Forcing~\cite{yi2025deep} and $\infty$-RoPE~\cite{yesiltepe2025infinity}). For long video generation, we adopt relative RoPE~\cite{liu2025rolling, yi2025deep, zhao2026relaxforcing}, where the RoPE indices of the global context are shifted to align with the immediately preceding history frames. For fair comparison, we also incorporate the Causal Forcing baseline with both global sink tokens and relative RoPE.

\section{User Study}
\label{app:user_study}

\begin{figure}[h]
    \centering

    \begin{subfigure}[t]{0.45\linewidth}
        \centering
        \includegraphics[width=\linewidth]{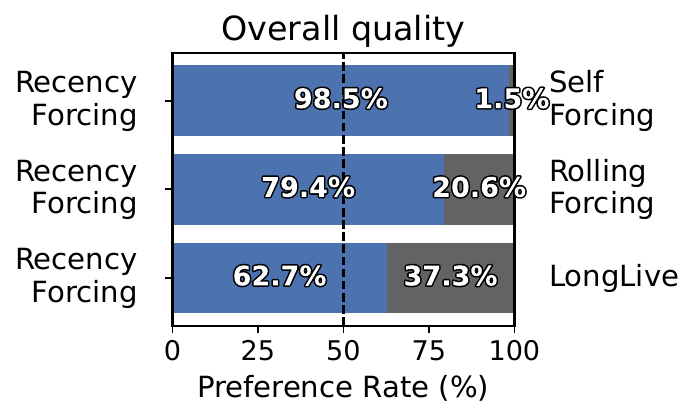}
        \caption{Overall quality.}
        \label{fig:user_overall}
    \end{subfigure}
    \begin{subfigure}[t]{0.45\linewidth}
        \centering
        \includegraphics[width=\linewidth]{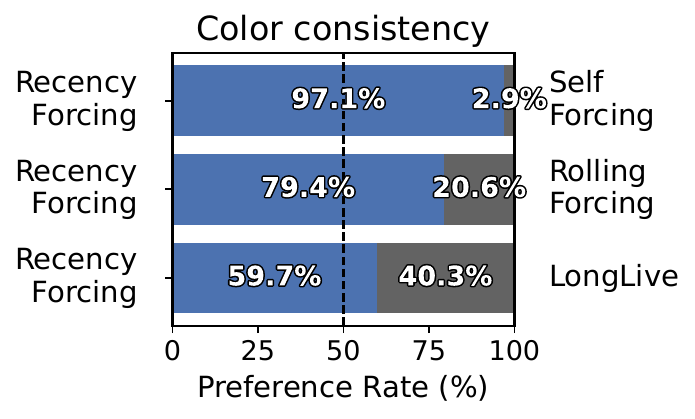}
        \caption{Color consistency.}
        \label{fig:user_color}
    \end{subfigure}

    \begin{subfigure}[t]{0.45\linewidth}
        \centering
        \includegraphics[width=\linewidth]{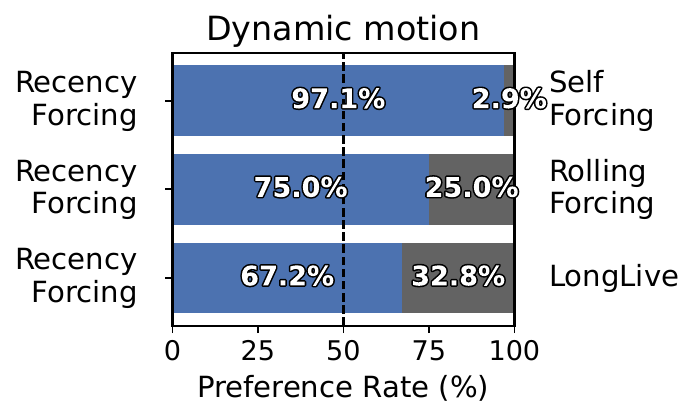}
        \caption{Dynamic motion.}
        \label{fig:user_motion}
    \end{subfigure}
    \begin{subfigure}[t]{0.45\linewidth}
        \centering
        \includegraphics[width=\linewidth]{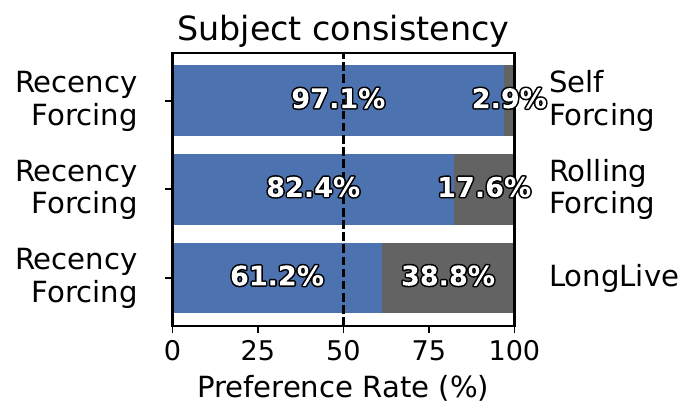}
        \caption{Subject consistency.}
        \label{fig:user_subject}
    \end{subfigure}

    \caption{
    \textbf{User preference study.}
    Preference rates of \textbf{Recency Forcing} against baseline methods across four evaluation aspects: \textbf{(a)} overall quality, \textbf{(b)} color consistency, \textbf{(c)} dynamic motion, and \textbf{(d)} subject consistency. Recency Forcing is consistently preferred in all settings. The largest margin is observed in \emph{dynamic motion}, highlighting improved long-range identity preservation beyond what is reflected by automatic metrics.
    }
    \label{fig:user_study}
\end{figure}

We conduct a human evaluation using a one-versus-other protocol.

\myheading{Setup.}
We conduct a user study with 23 participants and evaluate on randomly sampled 20 prompts. For each prompt, we generate videos from four methods: Self Forcing~\cite{Huang2025SelfFB}, Rolling Forcing~\cite{li2026rolling}, Longlive~\cite{yang2025longlive}, and our Recency Forcing. Each participant is assigned up to 20 pairwise comparisons randomly sampled from four methods.

\myheading{Protocol.}
We adopt a Two-Alternative Forced Choice (2AFC) protocol. Participants are shown two videos at a time and asked to select the better one according to the following criteria:
\begin{itemize}
    \item \textbf{Color consistency}: which video has more consistent colors across frames?
    \item \textbf{Dynamic motion}: which video show more natural and convincing motion dynamics?
    \item \textbf{Subject consistency}: which video keep the subject more consistent over time?
    \item \textbf{Overall quality}: Which video has better overall quality?
\end{itemize}

The evaluation is conducted via a web-based interface with randomized ordering to avoid bias.

\myheading{Statistics.}
In total, we collected over 1620 responses. Results are reported as preference percentages.

\myheading{Results.}
As shown in \cref{fig:user_study}, participants consistently prefer Recency Forcing across all four evaluation aspects. The largest margin appears in motion dynamics, suggesting that TRB’s suppression of distant-frame drift improves perceptual identity preservation --- an effect not fully captured by automatic metrics. These results further support our quantitative and qualitative findings.

\begin{figure}[t]
    \centering
    \includegraphics[width=\linewidth]{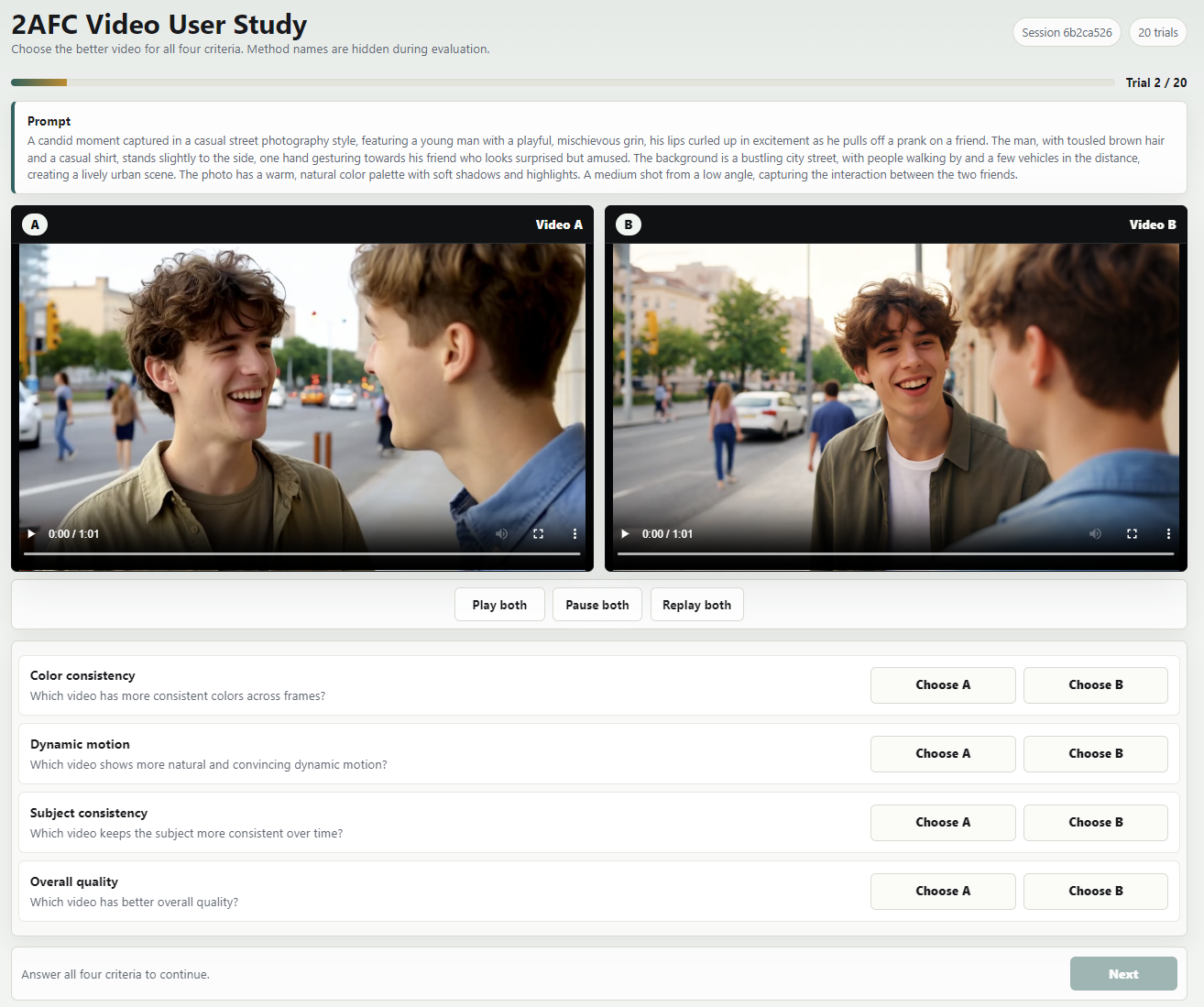}
    \caption{An example interface for our user preference study, where videos generated by different methods are displayed in a randomized
left/right arrangement.}
    \label{fig:user-study-ui}
\end{figure}

\section{Details of Biased Attention Reparameterization}

We provide the full derivation of \textbf{Biased Attention Reparameterization (BAR)}.

For a single query frame $q_i$, biased attention is defined as:
\begin{equation}
o_i =
\operatorname{Softmax}\left(
\frac{q_i K^\top}{\sqrt{d}} + B_{i,:}
\right)V.
\end{equation}

Expanding the softmax:
\begin{equation}
o_i =
\frac{
\sum_j
\exp\left(
\frac{q_i k_j^\top}{\sqrt{d}} + B_{i,j}
\right)
v_j
}{
\sum_j
\exp\left(
\frac{q_i k_j^\top}{\sqrt{d}} + B_{i,j}
\right)
}.
\end{equation}

Since $B_{i,j}$ depends only on position, define:
\begin{equation}
w_j = \exp(B_{i,j}) \in (0,1].
\end{equation}

Substituting:
\begin{equation}
o_i =
\frac{
\sum_j
\exp\left(
\frac{q_i k_j^\top}{\sqrt{d}}
\right)
w_j v_j
}{
\sum_j
\exp\left(
\frac{q_i k_j^\top}{\sqrt{d}}
\right)
w_j
}.
\end{equation}

We absorb $w_j$ into an augmented value:
\begin{equation}
\tilde{v}_j = w_j \cdot [v_j;1]
\in \mathbb{R}^{d+1},
\end{equation}
where $[;]$ denotes concatenation. Stacking all augmented values gives
$\tilde{V} \in \mathbb{R}^{N \times (d+1)}$.

The attention can then be rewritten as standard attention:
\begin{equation}
\tilde{o}_i =
\operatorname{Softmax}\left(
\frac{q_i K^\top}{\sqrt{d}}
\right)\tilde{V}.
\end{equation}

Finally, the original output is recovered by:
\begin{equation}
o_i =
\tilde{o}_i[{:}d] ,/, \tilde{o}_i[d].
\end{equation}

This reformulation is algebraically identical to the original biased attention with no approximation, enabling efficient implementation using standard FlashAttention kernels.

\section{Limitations}
\begin{figure}[t]
    \centering
    \includegraphics[width=\linewidth]{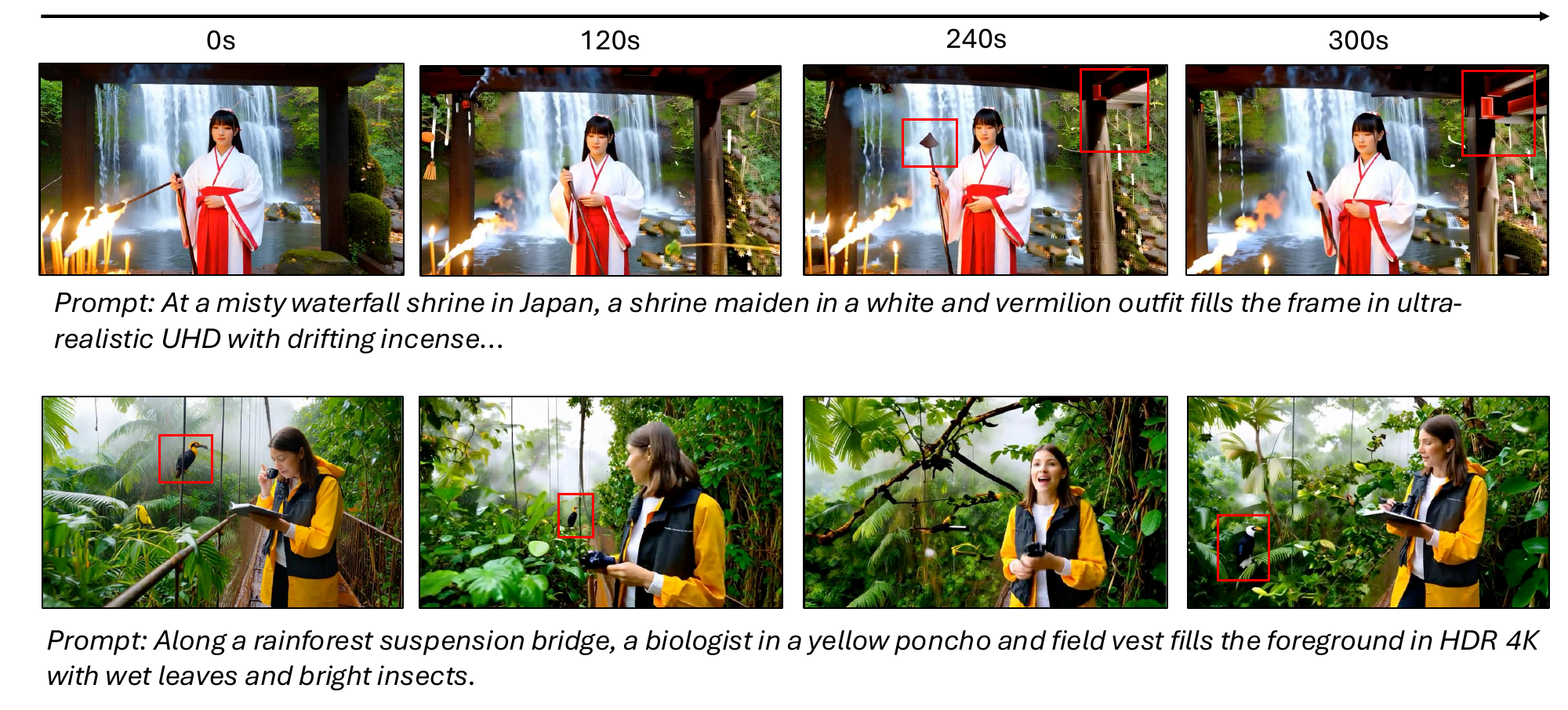}
    \caption{Failure cases of our method at extremely long horizons (5-minute videos). In the first-row example, the house's column gradually drifts and changes color after approximately 2 minutes. In the second-row example, the bird disappears as the video approaches the 4-minute mark, then reappears with inconsistent visual appearance compared to earlier frames.}
    \label{fig:limitation}
\end{figure}

Though our method can be applied to generate arbitrarily long videos, we observe degradation when the duration exceeds 4-5 minutes. As shown in \cref{fig:limitation}, the first-row sample exhibits visual inconsistency and gradual drift in the house column over time, while the bird in the second-row sample disappears and later reappears with inconsistent appearance. Generating infinitely long videos without quality degradation remains highly challenging, as previous methods typically demonstrate generation only up to limited durations (usually around 120 seconds).

\end{document}